# Hybridization of filter and wrapper approaches for the dimensionality reduction and classification of hyperspectral images

[1]Asma Elmaizi, [2]Maria Merzouqi, [3]Elkebir Sarhrouni, [4]Ahmed hammouch, [5]Chafik Nacir

Electrical Engineering Research Laboratory, ENSET Mohammed V University Rabat, Morocco

[1]asma.elmaizi@gmail.com, [2]merzouqimaria@gmail.com, [3]sarhrouni436@yahoo.fr, [4]hammouch_a@yahoo.com, [5]nacir_chafik@yahoo.fr

*Abstract*— The high dimensionality of hyperspectral images often imposes a heavy computational burden for image processing. Therefore, dimensionality reduction is often an essential step in order to remove the irrelevant, noisy and redundant bands. And consequently, increase the classification accuracy. However, identification of useful bands from hundreds or even thousands of related bands is a nontrivial task. This paper aims at identifying a small set of bands, for improving computational speed and prediction accuracy. Hence, we have proposed a hybrid algorithm through band selection for dimensionality reduction of hyperspectral images. The proposed approach combines mutual information gain (MIG), Minimum Redundancy Maximum Relevance (mRMR) and Error probability of Fano with Support Vector Machine Bands Elimination (SVM-PF). The proposed approach is compared to an effective reproduced filters approach based on mutual information. Experimental results on HSI AVIRIS 92AV3C have shown that the proposed approach outperforms the reproduced filters.

## I. Introduction

Hyperspectral sensors collect the images simultaneously in hundreds of contiguous spectral bands, with wavelengths ranging from visible to infrared [1]. The high dimensionality of spectral bands provides rich and detailed spectral information, which helps detecting targets and classifying materials with potentially higher accuracy. However, it brings challenges for hyperspectral image processing.

In fact, the measured bands are not necessarily all needed for an accurate discrimination and the use of the entire set of bands can lead to a poor classification model. This is due to the curse of dimensionality [8].

Dimensionality reduction based on band selection is an essential step for hyperspectral images processing. The development of bands selection has two major directions.

The filters work fast using a simple measurement [9], but its measurements results are not always satisfactory. On the other hand, the wrappers guarantee good results through examining learning results, but they are very slow when applied to HIS that contain hundreds of bands. Though the filters are very efficient in selecting bands, they are unstable when performing on wide bands sets. This research tries to incorporate the wrappers to deal with this problem.

The proposed algorithm is not a pure wrapper procedure, but rather a hybrid band selection model which utilizes both filter and wrapper methods. This paper introduces a new approach to the HSI reduction and suggests that the proposed approach can provide a better bands classification results by employing a three-stage selection algorithm and hybridizing the information gain (IG), the maximum relevance minimum redundancy(MRMR) filter (as filters method) and SVM-PF (Probability of Fano) as a wrapper method for addressing band selection problem.

The approach was evaluated using HSI AVIRIS 92AV3C illustrated on the figure 1 provided by the NASA [2].

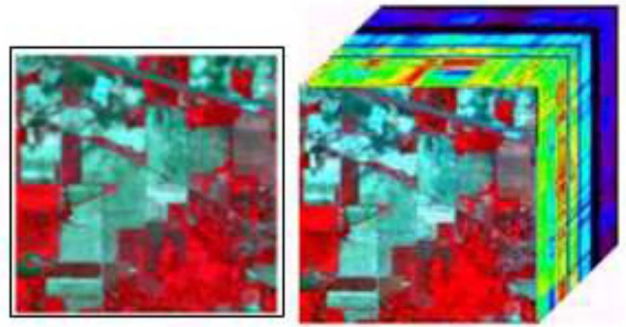

Fig 1. 3D cube of Indiana pines data set.

The AVIRIS 92AV3C (Airborne Visible Infrared Imaging Spectrometer) contains 220 images taken on the region "Indiana Pine" at "north-western Indiana", USA [2]. The 220 called bands are taken between 0:4 μm and 2:5 μm. Each band has 145 lines and 145 columns. The ground truth map is also provided (figure 2), but only 10366 pixels are labeled from 1 to 16. Each label indicates one from 16 classes. The zeros indicate pixels that are not classified yet.

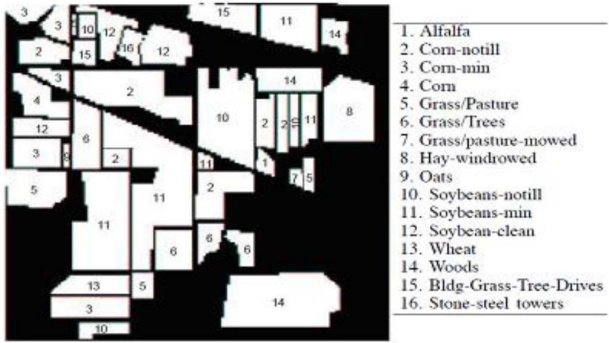

Fig 2. The Ground Truth map of AVIRIS 92AV3C

## II. THE MUTUAL INFORMATION AND THE PROBABILITY OF ERROR

This work proposes a method to reduce the dimensionality of the image while reassuring the good prediction within a minimum time. The algorithm developed is based on three steps, as a first step we computed the gain of mutual information, and in the next stages, we have combined two approaches which will be distinctly described in the next paragraph: the filter MRMR and the wrapper with error probability.

### A. First stage: Gain of mutual information

To quantify the amount of information contained by a random variable A, Shannon entropy, denoted by H(A) [10], is defined as:

$$H(C) = \sum_C p(C) \log_2 p(C) \quad (1)$$

Where p(C) is the probability density function (pdf) of C.

Guo [4] and Sarhrouni [5] used the mutual information formula shown below (2) in their work on the same image HSI AVIRIS 92AV3C [3]. Mutual information is widely used in the selection, in our case it was used for the reduction of the dimensionality of HSI by the selection of good bands by measuring the statistical dependence between two random variables C and B.

$$I(C, B) = \sum_{C,B} P(C, B) \log_2 \frac{P(C,B)}{P(C)p(B)} \quad (2)$$

### B. Second Stage: minimum redundancy –Maximum Relevance (mRMR).

Min redundancy and max relevance (mRMR) is a filter approach based on standard statistical measures, which is in our case: Mutual information. It consists in introducing selection procedures independently of the learning algorithm that will be developed and used thereafter. MRMR proposed by Peng et al. [15] comes to solve the redundancy issue between the selected bands. It tries to select the bands that have the highest value of mutual information and reduces redundancy using the evaluation function formula defined below (3). It is well known for its speed and flexibility.

$$mRMR = \max_{g_i \in G - S_{m-1}}[I(g_i; C) - \frac{1}{m-1} \sum_{g_j \in S_{m-1}} I(g_j; g_i)] \quad (3)$$

This expression is based on the conjoint information I (; c), consists of two parts: the first part serves to calculate it between the bands which have the high value of the mutual information and the ground truth C, then the second part is the conjoint information calculated between Each pair of bands. With the m-1 bands.

### C. Third Stage: the wrapper approach probability of Fano "PF"

In order to deal with the redundancy, fano [11] [16] has proposed the expressions below in the wrapper strategy. This demonstrates that the probability error and the mutual information vary contrarily. A low probability error means a high value of the mutual information for the selected band (is nearer to the ground truth map).

With: $\frac{H(C \backslash X) - 1}{Log_2 N_c} < P_e < \frac{H(C \backslash X)}{Log_2 N_c}$

$$\frac{H(C \backslash X) - 1}{Log_2 N_c} = \frac{H(C) - I(C, X) - 1}{Log_2 N_c} \quad (4)$$

Or :

$$P_e \leq \frac{H(C) - I(C, X) - 1}{Log_2 N_c} = \frac{H(C \backslash X) - 1}{Log_2 N_c}$$

To compute the probability error, it is necessary to compute the conditional entropy between the selected images (X) and the ground truth (C) as already presented in the formula above H (C / X). Represent the number of classes.

## III. THE PROPOSED APPROACH

This paper aims to overcome the hyperspectral images challenges by selecting the most important bands containing relevant information.

As already mentioned at the beginning, to get good results, a hybrid method is proposed for using the benefits of each stage.

The process followed to select bands:

**The proposed Algorithm:**

1. A founded order of the bands on a descending scale according to their values of the mutual information is made. According to Their rankings we eliminate the irrelevant ones.

2. We calculate the filter mRMR evaluation function defined on the equation (3) to eliminate all the redundant and irrelevant bands.

3. We choose the first band provided by the previous step to build an approximation of the ground truth (GT_est).

4. The fano wrapper serves for increased precision by eliminating the redundant one. On each-iteration we calculate the probability of fano, using the equation (4). For the candidate bands to be taken. It will be necessary to reduce the last values of the retained Pe by adding a given threshold.

5. As a final step: the construction of a new ground truth GT (GT_est) is done by classification and it based on the bands considered as the best and the SVM[13].

To achieve our objective, our idea is based on the following process. We start by computing the gain of the mutual information between each pair of the field truth pair and every band in order to select those with the highest value of Mi. However, the problem is still remaining because of the redundancy and correlated bands obtained at the end of the first stage. To solve the previous issue, we have used the filter MRMR in the second stage, to benefit from its selection effectiveness. MRMR filter hold the relevant band and eliminates the redundant and the irrelevant as well. This involves minimizing the computing time which is a way out to avoid the stress that we will get in the third stage. As we are greedy to get the best of the best result, we add the Fano wrapper that can reach a better result than a filter does. The Wrapper Fano selects bands that minimize the error probability with high accuracy from the retained bands in the previous stage. So, this Algorithm permits to achieve a good result with high accuracy in a short delay.

IV. EXPERIMENT, RESULTS AND ANALYSIS

To test the performance of the methods, 50% labeled pixels are randomly selected to be used in training, and the other 50% will be used for the classification testing [12]. The classifier used is SVM [13] [14]. The average classification accuracy is used as a performance measure.

A. Results

The following table 1 presents the results obtained using the proposed method based on the hybrid approach MIG, MRMR and SVM-PF. The classification accuracy is calculated over the whole size of the selected subset, starting from 2 bands up to 80 bands for different threshold values.

| | Th | -0,02 | -0,005 | -0,0035 | 0 |
|---|---|---|---|---|---|
| Number of retained Bands | 2 | 36,22 | 36,22 | 36,22 | 36,22 |
| | 3 | 38,07 | 38,07 | 38,07 | 38,07 |
| | 4 | 47,54 | 47,54 | 47,54 | 47,54 |
| | 12 | 60,02 | 62,97 | 62,98 | 57,76 |
| | 14 | 60,24 | 64,15 | 64,66 | 58,50 |
| | 18 | 66,65 | 66,73 | 70,41 | 72,59 |
| | 20 | 69,57 | 76,50 | 78,49 | 75,59 |
| | 25 | 77,58 | 78,55 | 79,72 | |
| | 35 | 77,87 | 79,70 | 80,65 | |
| | 36 | 78,71 | 81,12 | 81,49 | |
| | 40 | 79,12 | 81,63 | 81,86 | |
| | 45 | 80,87 | 82,82 | 82,84 | |
| | 50 | 82,12 | 84,12 | 84,28 | |
| | 53 | 8284 | 85,70 | 85,54 | |
| | 60 | 83,11 | 8588 | 86,42 | |
| | 70 | 85,27 | 86,85 | 86,91 | |
| | 75 | 86,42 | | | |
| | 80 | 86,60 | | | |

The accuracy (%) of classification for numerous thresholds

TABLE I. Results of the hybrid algorithm proposed (IG-MI-PF).

The table II gives the classification results using an information gain filters selection by mutual information an SVM classifier [13] [14].

| | 5 | 10 | 15 | 20 | 25 | 30 | 35 |
|---|---|---|---|---|---|---|---|
| IG | 51,82 | 55,43 | 57,65 | 63,08 | 66,12 | 73,54 | 76,06 |
| | 40 | 45 | 50 | 55 | 60 | 70 | 80 |
| | 78,96 | 80,58 | 81,63 | 82,06 | 82,74 | 86,58 | 86,89 |

The classification accuracy (%)

TABLE II: Results of the information gain algorit*hm*

The result of the algorithm based on the mutual information [5] using the classifier SVM and a given thresholds ("Th" in the aim to control the redundancy), will be shown on the table III below:

| Th | \multicolumn{4}{c}{The rate of classification for different thresholds} |
| --- | --- | --- | --- | --- |
| | -0,02 | -0,0050 | -0,0035 | -0,000 |
| 2 | 47,44 | 47,44 | 47,44 | 47,44 |
| 3 | 47,87 | 47,87 | 47,87 | **48,92** |
| 4 | 49,31 | 49,31 | 49,31 | |
| 12 | 56,30 | 56,30 | 60,76 | |
| 14 | 57,00 | 57,00 | 61,80 | |
| 18 | 59,09 | 59,09 | **63,00** | |
| 20 | 63,08 | 63,08 | | |
| 25 | 66,12 | 64,89 | | |
| 35 | 76,06 | 75,59 | | |
| 36 | 76,49 | **76,19** | | |
| 40 | 78,96 | | | |
| 45 | 80,85 | | | |
| 50 | 81,63 | | | |
| 53 | 82,27 | | | |
| 60 | 82,74 | | | |
| 70 | 86,95 | | | |
| 75 | 86,81 | | | |
| 80 | **87,28** | | | |

TABLE III. Results of filter (MI) [5]: Redundancy check for several given thresholds (Th)

*B. Analysis and discussion*

The above table (I) shows the strength of our proposed approach compared with information selection filter table (II) and the mutual information, redundancy elimination filter [5].

The analysis of the previous tables allowed us to take out the following results:

- As shown on the Table (III), the chosen negative thresholds reflect negatively on the classification rate and the redundancy control.

- At the same time, table (I) shows that the proposed approach achieves the highest accuracy and outperforms the redundancy elimination algorithm (table III) using the same negative thresholds for the proposed algorithm.

- It achieves 70% classification accuracy with 18 bands, which is higher than the MI filter (Table III) by 7, 41% for (Th=-0, 0035), and higher than the IG filter by 7% (Table II) for the same retained bands number.

- For relatively large thresholds (-0.0035, 0) filters (Table III) selection is hard, there is no redundancy. For example, with Th=0, just three bands are retained for the filters (Table III). However, the proposed algorithm increases the classification rate and we achieved 75% with 20 retained band for the same Th=0.

- For the average thresholds (-0.005) with the introduced method, we had an important result with a minimal of the bands. It's increased the classification rate from 76 to 81 with the same number of bands selected, which confirms that the proposed approach (Table I) surpasses the two filters on the accuracy of the classification and the number of allowed bands retained

- It achieves 76% classification accuracy with 20 bands, which is higher than the MI filter (Table III) by 13% for (Th=-0,005), and higher than the IG filter by 13% (Table III) for the same threshold.

- For negative thresholds (Th = -0.02), there will be more tolerance for the redundant thing which explains the high numbers of bands for the same classification rate. According to the results obtained (for 53 bands), the proposed method is the most robust of the others (table II and III).

*C. Partial conclusion*

We can conclude that the proposed algorithm uses a hybrid approach based on three stages (IG, mRMR and PF), using mutual information, without using a pure wrapper that could slows down the selection process. Its major advantage is the redundancy, relevance control and classification accuracy increasing compared to a pure filter as shown on the table (I, II, III). We benefit from the function used on each stage and we achieve higher accuracy due to the including of the wrapper stage. Hence, the hybrid method is a perspective of the filter and wrapper approach which has improved the results in a short time.

V. CONCLUSION

This paper presents a new bands-selection filter method based on a hybrid approach to face the hyperspectral image classification challenges by reducing their dimensionality.
In this paper, a new algorithm was introduced based on three- stage processes, each one with different role. In the first stage, IG was used to identify a candidate band in order to choose the relevant bands which has the highest value of IG. The mRMR filter had been applied in the second stage in order to benefit from the redundancy minimizing and selecting effective bands, from the set recuperated in the first stage. As a final stage, SVM-PF wrapper approach was applied to make sure that we achieve a high accuracy. The approach is assessing the classification accuracy by SVM-RBF classifier in the last stage. We take advantages of both the filter and the wrapper. It is not as fast as a pure filter, but it can achieve a better result than a filter does. Most importantly, the computational time and complexity can be reduced in comparison to a pure wrapper. The results had illustrated that the proposed method is very effective and practical having great potential for bands selection in a short delay with a high classification rate compared to the one presented by a filters or pure wrapper.